%% file: acl_latex.tex
\pdfoutput=1

\documentclass[11pt]{article}

\usepackage[preprint]{acl}

\usepackage{times}
\usepackage{latexsym}
\usepackage{algorithm}
\usepackage{algpseudocode}
\usepackage{amsmath}
\usepackage{multirow}
\usepackage{array}
\usepackage{multirow}
\usepackage{graphicx}
\usepackage{colortbl}
\usepackage{amssymb}
\usepackage{pifont} 
\usepackage{booktabs}
\usepackage{subcaption}  
\usepackage{caption}  

\usepackage[T1]{fontenc}

\usepackage[utf8]{inputenc}

\usepackage{microtype}

\usepackage{inconsolata}

\usepackage{graphicx}

\title{Sens-Merging: Sensitivity-Guided Parameter Balancing for Merging \\ Large Language Models}

\author{Shuqi Liu$^{1,2}$, Han Wu$^{2,}$$^\dagger$, Bowei He$^{1}$, Xiongwei Han$^2$, Mingxuan Yuan$^2$, Linqi Song$^{1,}$$^\dagger$\\
$^{1}$ Department of Computer Science, City University of Hong Kong\\
$^{2}$ Huawei Noah's Ark Lab\\
\texttt{\{shuqiliu4-c, boweihe2-c\}@my.cityu.edu.hk}\\
\texttt{wu.han1@huawei.com}\\
\texttt{linqi.song@cityu.edu.hk}
}

\begin{document}
\maketitle
\begin{abstract}
Recent advances in large language models have led to numerous task-specialized fine-tuned variants, creating a need for efficient model merging techniques that preserve specialized capabilities while avoiding costly retraining. While existing task vector-based merging methods show promise, they typically apply uniform coefficients across all parameters, overlooking varying parameter importance both within and across tasks. We present Sens-Merging, a sensitivity-guided coefficient adjustment method that enhances existing model merging techniques by operating at both task-specific and cross-task levels. Our method analyzes parameter sensitivity within individual tasks and evaluates cross-task transferability to determine optimal merging coefficients. Extensive experiments on Mistral 7B and LLaMA2-7B/13B models demonstrate that Sens-Merging significantly improves performance across general knowledge, mathematical reasoning, and code generation tasks. Notably, when combined with existing merging techniques, our method enables merged models to outperform specialized fine-tuned models, particularly in code generation tasks. Our findings reveal important trade-offs between task-specific and cross-task scalings, providing insights for future model merging strategies.
\end{abstract}

{
\let\thefootnote\relax\footnotetext{
$^\dagger$Corresponding author.}
}


\begin{figure}
    \centering
    \includegraphics[width=1.0\linewidth]{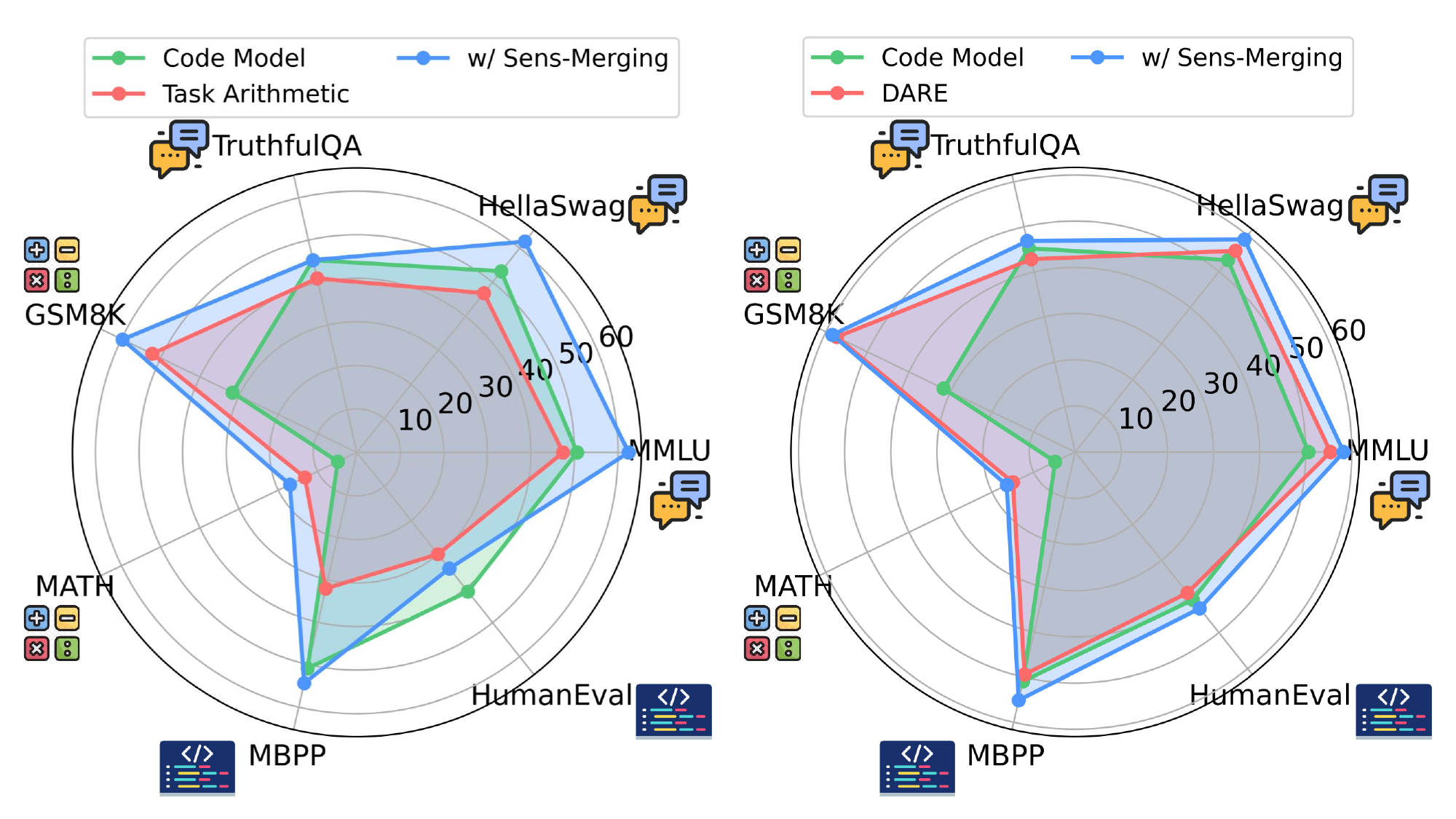}
    \caption{Sens-Merging functions as a plug-and-play enhancement to existing task vector-based merging techniques. Notably, when integrated with DARE, it surpasses even specialized code models in code generation.}
    \label{fig:radar_combined}
    \vspace{-0.3cm}
\end{figure}

\section{Introduction}

The rapid advancement of large language models has significantly enhanced performance across a diverse range of tasks \cite{touvron2023llama, zhao2023survey}. As these models continue to be fine-tuned for specialized domains, the necessity to merge these specialized models into a unified framework becomes increasingly critical \citep{merge_survey, mergeKit}. While multi-task learning has been proposed as a solution, it incurs substantial training costs and requires simultaneous access to data and labels for all tasks \cite{sanh2022multitask, fifty2021efficiently}. Recently, researchers have developed parameter-level model merging methods that not only comply with data privacy regulations but also improve efficiency by eliminating the need for retraining \cite{ties_merging, dare}.

In the context of model merging, task vectors \cite{ilharco2023editingmodelstaskarithmetic} have emerged as a powerful component for representing task-specific capabilities. These vectors, defined as the differences between parameter values before and after fine-tuning, enable effective integration of specialized knowledge from different models. 
While task vector-based merging methods \cite{ties_merging, dare} have shown promising results, their reliance on uniform coefficients for each task and parameter limits their potential effectiveness. 
This uniformity implies that every task and every parameter is treated with equal importance during the merging process. Consequently, it overlooks the fact that parameters within each layer demonstrate varying levels of importance for specific tasks, and parameters from different tasks contribute distinctly during the merging process.

To address these challenges, we propose Sens-Merging, a sensitivity-guided merging coefficient adjustment method that functions as a plug-and-play enhancement to existing task vector-based merging techniques. Our method operates at two levels: within individual tasks and across different tasks, allowing for fine-grained control over parameter importance. 
Within each task-specific model, we perform parameter sensitivity analysis to highlight critical layers that significantly impact performance. Concurrently, across different tasks, we conduct task sensitivity analysis to prioritize models that enhance the performance of others. By combining these two factors, we derive the final merging coefficients, which are then applied to merge the corresponding layers.
Figure \ref{fig:radar_combined} highlights how Sens-Merging enhances existing task-vector techniques like Task Arithmetic \cite{task_arithmetic} and DARE \cite{dare}. Notably, when combined with DARE method, Sens-Merging enables merged models to outperform specialized fine-tuned models, particularly in code generation tasks.

To empirically demonstrate the effectiveness of Sens-Merging, we conduct extensive experiments by combining it with existing model merging approaches. We merged three widely adopted fine-tuned models—specializing in general knowledge (Chat), mathematical reasoning (Math), and code generation (Code)—derived from the LLaMA2-7B/13B and Mistral 7B families. 
The integration of our Sens-Merging not only improves baseline merging performance but enables merged models to surpass individual fine-tuned models. 
Notably, when merging Code model with Math and Chat models using Sens-Merging, it achieves superior performance on coding tasks compared to code-specific fine-tuning alone. These results indicate that model merging can effectively address the challenges of training a single model for complex tasks by integrating the specialized capabilities of multiple fine-tuned models.

To sum up, our contributions include:
(1) We propose a novel model merging coefficient determination method based on both task-specific and cross-task sensitivity analysis.
(2) Through comprehensive evaluations, 
we validate that our proposed method enhances model merging performance across various domains.
(3) We empirically demonstrate that different task-specific models contribute unequally to model merging, and parameter importance varies across different layers within each model.
(4) We validate that each scaling approach presents distinct trade-offs: task-specific scaling excels in specialized domains like mathematics but offers limited general benefits, while cross-task scaling achieves broader performance gains at the cost of peak task-specialized performance.

\begin{figure*}[tbp]
    \centering
    \includegraphics[width=0.95\textwidth]{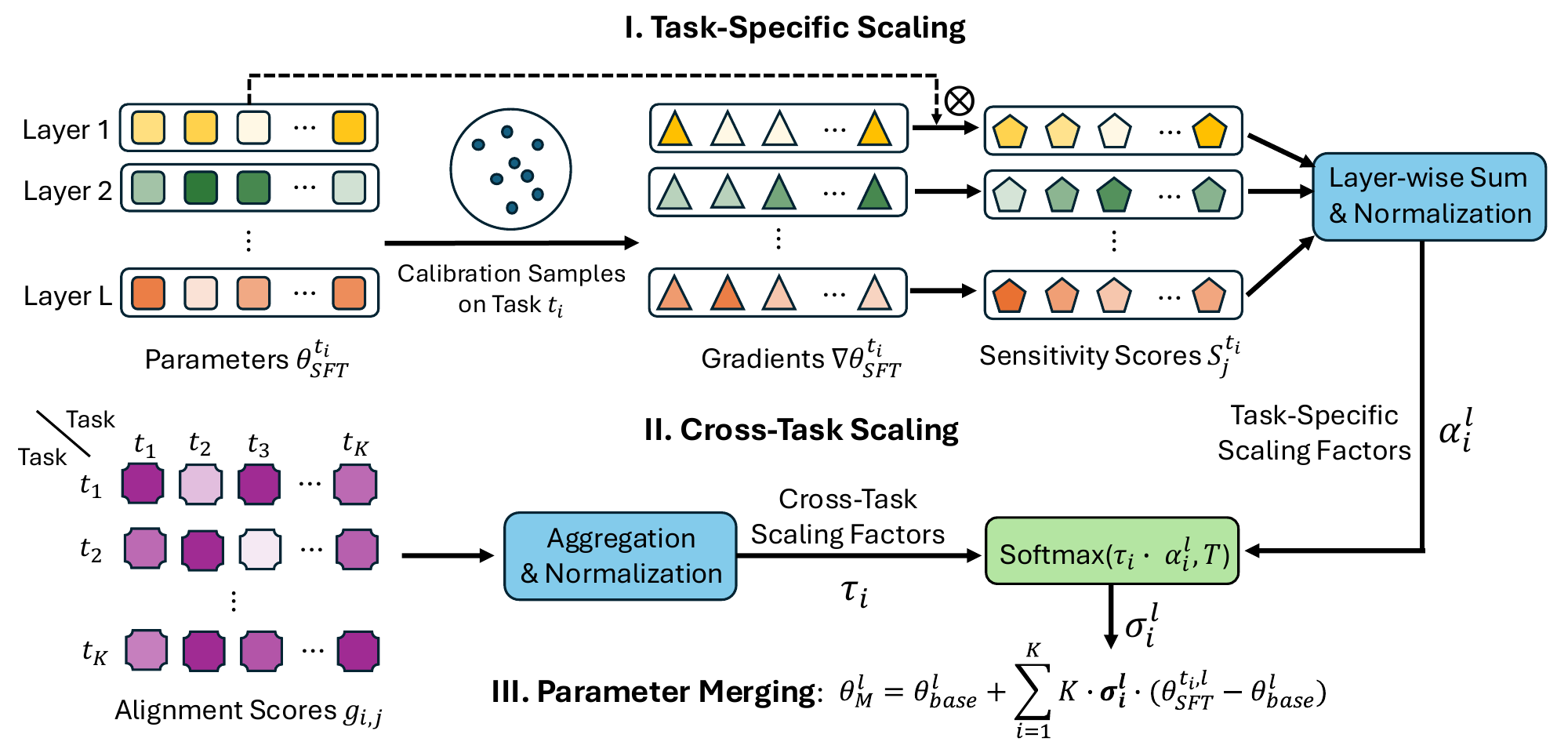}
    \caption{Overall framework of our Sens-Merging method. Sens-Merging adjusts layer-wise scaling coefficients for task-specialized fine-tuned models through two mechanisms: task-specific scaling and cross-task scaling.
    }

    \label{fig: framework}
    \vspace{-0.3cm}
\end{figure*}

\input{latex/related_work}

\section{Methodology}
Our Sens-Merging method combines two levels of sensitivity analysis: layer-wise analysis within individual models and cross-task analysis across different models to achieve a balanced parameter distribution. For layer-wise analysis, we compute sensitivity scores using gradient information from calibration datasets. For cross-task analysis, we evaluate model alignment through logit comparison. These two components determine the final merging coefficients used to merge corresponding layers into a unified model, as shown in Figure XX.

\subsection{Preliminary}
Considering \( K \) task-specialized fine-tuned models \( \{\theta^{t_1}_{\text{SFT}}, \ldots, \theta^{t_K}_{\text{SFT}}\} \) derived from a common pre-trained backbone \( \theta_{\text{PRE}} \), model merging aims to merge them into a single model $\theta_M$ that can effectively handle all tasks simultaneously. 
The task-specific capabilities of each fine-tuned model are captured by task vectors, defined as the difference between the fine-tuned parameters and the pre-trained backbone:
\[
\delta_{t_k} = \theta^{t_k}_{\text{SFT}} - \theta_{\text{PRE}}, \quad \text{for } k \in \{1, \ldots, K\}.
\]
Task vector-based merging aggregates these task vectors to construct a single, static merged model:
\[
\theta_{\text{M}} = \theta_{\text{PRE}} + \sum_{k=1}^{K} \lambda * \delta_{t_k}.
\]
where the coefficient $\lambda$ represents the importance of each merged task vector.

\subsection{Task-Specific Scaling}

To accurately balance the parameters within individual task models, we conduct layer-wise sensitivity analysis by measuring each layer's contribution to model performance through aggregating parameter sensitivities within that layer.

\paragraph{Parameter Sensitivity.}
We define parameter sensitivity as the change in loss when setting that parameter to zero. A parameter is considered highly sensitive if zeroing it results in a significant loss increase.
For a fine-tuned model with parameters $\theta^{t_i}_{\text{SFT}} = [\theta_1, ..., \theta_{N}]$, where $N$ represents the total number of parameters, the $j$-th parameter can be expressed as $\theta^{t_i}_{j} = [0, ..., \theta_j, ..., 0]$. With gradients of the loss relative to $\theta^{t_i}_{\text{SFT}}$ represented as $\nabla_{\theta^{t_i}_{\text{SFT}}} L$, the sensitivity of the $j$-th parameter for a specific sample $x_k$ from task $t_i$ is determined as:
\begin{equation}
S_{j,k}^{t_i} = |(\theta^{t_i}_{j})^\top \nabla_{\theta^{t_i}_{\text{SFT}}} L(x_k)|
\end{equation}
The rationale behind this sensitivity definition stems from the first-order Taylor expansion of $L(x_k)$ relative to $\theta_j$. In essence, $S_{j,k}^{t_i}$ provides an approximation for how the loss might change in the absence of $\theta_j$:
\begin{equation}
(\theta^{t_i}_{j})^\top \nabla_{\theta^{t_i}_{\text{SFT}}} L(x_k) \approx L(\theta^{t_i}_{\text{SFT}}) - L(\theta^{t_i}_{\text{SFT}} - \theta^{t_i}_{j})
\end{equation}
To estimate the parameter sensitivity $S_{j}^{t_i}$ for task $t_i$,
we randomly sample $m$ instances from the task training set as calibration samples. 
The final sensitivity score $S_j^{t_i}$ aggregates the individual sensitivities across all sampled instances:
$S_{j}^{t_i} = \sum_{k=1}^m S_{j,k}^{t_i}$.


\paragraph{Layer-Wise Sensitivity and Normalization.}

The layer-wise sensitivity \( s_i^l \) is then calculated by summing the sensitivities of all parameters within each layer, thereby reflecting each layer's overall contribution to the model's performance. To allow for meaningful comparisons of these importance scores across different models, we apply \( L_2 \) normalization to the sensitivities of all layers.
Consequently, the task-specific sensitivity scaling factors \( \alpha_i^l \) are defined as:
\begin{equation}
    s_i^l = \sum_{j \in \mathcal{P}_l} S_{j}^{t_i}, \quad \alpha_i^l = \frac{s_i^l}{\|\mathbf{s}_i\|_2}
\end{equation}
where \( \mathcal{P}_l \) denotes the set of parameters in layer \( l \), and \( L \) is the total number of layers in the model. 

\subsection{Cross-Task Scaling}


While task-specific sensitivity focuses on the importance of layers within individual tasks, it is equally essential to evaluate how each task-specific model influences other tasks during the merging process. Cross-task sensitivity captures the interdependencies and shared representations between different tasks, ensuring that the merged model benefits from common features and decision-making processes.

The measurement of cross-task influence begins with evaluating logits alignment between different task-specific models. Specifically, for calibration samples from task $t_j$, we compute the alignment score between model $\theta^{t_i}_{\text{SFT}}$ and the expert model for task $t_j$, $\theta^{t_j}_{\text{SFT}}$, using the $L_2$ distance between their output logits:
\begin{equation}
g_{i,j} = ||f_{\theta^{t_i}_{\text{SFT}}}(x_k^j) - f_{\theta^{t_j}_{\text{SFT}}}(x_k^j)||_{2}
\end{equation}
where $f_{\theta}(x)$ denotes the output logits of model $\theta$ for input $x$, and $||\cdot||_{2}$ represents $L_{2}$ distance. This alignment score quantifies how closely the predictions of model $\theta^{t_i}_{\text{SFT}}$ match those of the expert model for task $\theta^{t_j}_{\text{SFT}}$, providing insight into the degree of shared knowledge and representational similarity between tasks.
To obtain a comprehensive measure of cross-task sensitivity for a specific task model $\theta^{t_i}_{\text{SFT}}$, we aggregate the alignment scores across all other tasks. This aggregation process involves computing the normalized alignment:
\begin{equation}
    \tau_i = \sum_{i=1, i\neq j}^{K} g_{i,j}, \quad \tau_i = \frac{\tau_i}{\|\boldsymbol{\tau}\|_1}
\end{equation}
The resulting cross-task scaling factor $\tau_i$ serves as a crucial metric that quantifies model $\theta^{t_i}_{\text{SFT}}$'s ability to transfer knowledge across tasks. Higher values of $\tau_i$ indicate superior cross-task generalization capabilities, suggesting that the model has learned robust representations that are valuable across multiple tasks. Conversely, lower values of $\tau_i$ reflect greater task-specific specialization, indicating that the model's features are more narrowly focused on its primary task. 

\subsection{Integration with Merging Methods}

Our Sens-Merging method combines task-specific scaling factor $\alpha_i^l$ and the cross-task scaling factor $\tau_i$ into a plug-and-play module, which can be seamlessly integrated with existing task vector-based model merging methods. 
To effectively combine these sensitivity factors, we employ a two-step process. First, we multiply the task-specific scaling factor $\alpha_i^l$ with the cross-task scaling factor $\tau_i$ to capture both task-specific and cross-task importance. Then, we apply a softmax function with temperature T to normalize these products and obtain the final scaling coefficients:
\begin{equation}
    \sigma_i^l = \text{Softmax} (\tau_i \cdot \alpha_i^l, T)
\end{equation}

The final step involves computing the merged model parameters $\theta_M^l$ for each layer $l$. We start with the base model parameters $\theta_{\text{base}}^l$ and incorporate weighted contributions from all $K$ fine-tuned models. The contribution of each task-specific model is scaled by its normalized coefficient $\sigma_i^l$ and multiplied by $K$ to preserve the magnitude of updates:
\begin{equation}
\theta_M^l = \theta_{\text{base}}^l + \sum_{i=1}^K K \cdot \sigma_i^l \cdot (\theta_{\text{SFT}}^{t_i,l} - \theta_{\text{base}}^l)
\end{equation}


\section{Experiments}


\paragraph{Baselines.} 
We evaluate the effectiveness of our Sens-Merging method by comparing it against both individual task-specific models and several established model-merging techniques, including Task Arithmetic, Ties-Merging, and DARE-Merging. Task Arithmetic \citep{task_arithmetic} enhances the merging process by introducing task vectors, suggesting that simple arithmetic operations on these vectors can effectively edit models and produce a merged model. Building on the concept of task vectors, both DARE \citep{dare} and Ties \citep{ties_merging} employ pruning-then-scaling methods to merge task vectors, based on the assumption that not all parameters contribute equally to the final performance.  

\paragraph{Hyperparameters.}
Both baselines and our Sens-Merging enhanced baselines use the same hyperparameters for fair comparison. 
For Task Arithmetic, we use a default scaling coefficient of $\lambda = 1$, which maintains the original magnitude of task vector when adding the pretrained backbone. 
However, the DARE method has been observed to be more sensitive to variations in both the scaling coefficient \(\lambda\) and the drop rate parameter \(r\). 
To achieve a balanced performance, we set the scaling coefficient to \(\lambda = 0.5\) and establish a default drop rate of \(r = 0.5\) for DARE.
Similarly, for Ties-Merging, which requires the specification of a masking ratio, we set the default mask ratio to \(r = 0.7\) across all experiments. 


\paragraph{Benchmark.} 
Our experimental evaluation encompasses three families of models: LLaMA-2 7B series \cite{touvron2023llama}, Mistral 7B series \cite{jiang2023mistral}, and LLaMA-2 13B series \cite{touvron2023llama}, each specialized in distinct domains: general knowledge, mathematical reasoning, and code generation. For comprehensive evaluation, we utilize seven benchmark datasets spanning three key domains: MMLU \cite{hendrycks2020measuring}, HellaSwag \cite{zellers2019hellaswag} and TruthfulQA \cite{lin2022truthfulqa} for assessing general knowledge and reasoning capabilities; GSM8K \cite{cobbe2021training} and MATH \cite{hendrycks2021measuring} for testing mathematical reasoning proficiency; and HumanEval \cite{chen2021evaluating} and MBPP \cite{austin2021program} for evaluating code generation abilities. To ensure consistent and unbiased assessment, model performance is evaluated using zero-shot accuracy, with pass@1 rate specifically measuring code generation correctness.

\begin{table*}[]
\caption{Performance evaluation of merged LLaMA2-7B Models (Chat, Math, Code) across 7 task-specific datasets}
\resizebox{\textwidth}{!}{%
\begin{tabular}{lcccclcclccc} 
\toprule
& & \multicolumn{3}{c}{General Knowledge} &
   &
  \multicolumn{2}{c}{Mathemetical Reasoning} &
   &
  \multicolumn{2}{c}{Code Generation} &
  \multicolumn{1}{c}{} \\ \cline{3-5} \cline{7-8} \cline{10-11} 
\multirow{-2}{*}{Method} &
  \multirow{-2}{*}{\begin{tabular}[c]{@{}c@{}}Use\\ Sens\end{tabular}} &
  \multicolumn{1}{c}{MMLU} &
  \multicolumn{1}{c}{HellaSwag} &
  \multicolumn{1}{c}{TruthfulQA} &
   &
  \multicolumn{1}{c}{GSM8K} &
  \multicolumn{1}{c}{MATH} &
   &
  \multicolumn{1}{c}{MBPP} &
  \multicolumn{1}{c}{HumanEval} &
  \multicolumn{1}{c}{\multirow{-2}{*}{Average}} \\
  \hline
Chat & & 46.38 & 57.79 & 45.17 &  & 23.43 & 4.86 & & 0.3 & 0.6 & 25.50 \\
Math & & 40.05 & 56.30 & 32.56 &  & 48.60 & 8.50 &  & 21.8 & 12.8 & 31.52\\
Code & \multirow{-3}{*}{\textbackslash{}} & 40.76 & 57.87 & 33.17 &  & 7.13 & 3.62 &  & 26.8 & 5.5 & 24.98\\
\hline
 & \ding{55} & 41.50 & 49.63 & 37.45 &  & \underline{47.34} & 6.46 &  & 13.5 & 7.3 & 29.03\\
\multirow{-2}{*}{Task Arithmetic} &
  \checkmark &
  \cellcolor[HTML]{EFEFEF}46.12 &
  \cellcolor[HTML]{EFEFEF}\textbf{59.10} &
  \cellcolor[HTML]{EFEFEF}36.84 &
  \cellcolor[HTML]{EFEFEF} &
  \cellcolor[HTML]{EFEFEF}42.29 &
  \cellcolor[HTML]{EFEFEF}7.12 &
  \cellcolor[HTML]{EFEFEF}  &
  \cellcolor[HTML]{EFEFEF}\textbf{33.1} &
  \cellcolor[HTML]{EFEFEF}\underline{18.9} & 
  \cellcolor[HTML]{EFEFEF}34.78 \\
 & \ding{55} & 45.75 & 56.63 & \underline{39.89} &  & 46.93 & \underline{7.74} &  & 29.1 & 17.1 & 34.73 \\
\multirow{-2}{*}{Ties-Merging} &
  \checkmark &
  \cellcolor[HTML]{EFEFEF}46.03 &
  \cellcolor[HTML]{EFEFEF}56.87 &
  \cellcolor[HTML]{EFEFEF}\textbf{40.02} &
  \cellcolor[HTML]{EFEFEF} &
  \cellcolor[HTML]{EFEFEF}\textbf{47.69} &
  \cellcolor[HTML]{EFEFEF}\textbf{7.80} &
  \cellcolor[HTML]{EFEFEF} &
  \cellcolor[HTML]{EFEFEF}29.8 &
  \cellcolor[HTML]{EFEFEF}17.7 &
  \cellcolor[HTML]{EFEFEF}\textbf{35.13}\\
     & \ding{55}  & \underline{46.78} & 57.57 & 38.19 &  & 44.05 & 6.96 &  & 31.6 & \underline{18.9} & \underline{34.86}\\
\multirow{-2}{*}{DARE} & 
  \checkmark &
  \cellcolor[HTML]{EFEFEF}\textbf{46.81} &
  \cellcolor[HTML]{EFEFEF}\underline{58.24} &
  \cellcolor[HTML]{EFEFEF}37.33 &
  \cellcolor[HTML]{EFEFEF} &
  \cellcolor[HTML]{EFEFEF}44.73 &
  \cellcolor[HTML]{EFEFEF}6.98 &
  \cellcolor[HTML]{EFEFEF} &
  \cellcolor[HTML]{EFEFEF}\underline{32.3} &
  \cellcolor[HTML]{EFEFEF}\textbf{19.5} & 
  \cellcolor[HTML]{EFEFEF}\textbf{35.13} \\
  \bottomrule
\end{tabular}%
}
\label{tab:llama2-7b}
\end{table*}

\begin{table*}[]
\caption{Performance evaluation of merged Mistral 7B Models (Chat, Math, Code) across 7 task-specific datasets}
\resizebox{\textwidth}{!}{%
\begin{tabular}{lcccclcclccc}
\toprule
& & \multicolumn{3}{c}{General Knowledge} &
   &
  \multicolumn{2}{c}{Mathemetical Reasoning} &
   &
  \multicolumn{2}{c}{Code Generation} &
  \multicolumn{1}{c}{} \\ \cline{3-5} \cline{7-8} \cline{10-11} 
\multirow{-2}{*}{Method} &
  \multirow{-2}{*}{\begin{tabular}[c]{@{}c@{}}Use\\ Sens\end{tabular}} &
  \multicolumn{1}{c}{MMLU} &
  \multicolumn{1}{c}{HellaSwag} &
  \multicolumn{1}{c}{TruthfulQA} &
   &
  \multicolumn{1}{c}{GSM8K} &
  \multicolumn{1}{c}{MATH} &
   &
  \multicolumn{1}{c}{MBPP} &
  \multicolumn{1}{c}{HumanEval} &
  \multicolumn{1}{c}{\multirow{-2}{*}{Average}} \\
  \hline
Chat & & 59.05 & 65.97 & 55.69 &  & 42.53 & 9.16 & & 49.6 & 42.7 & 46.37 \\
Math & & 60.77 & 58.68 & 44.68 &  & 63.38 & 22.74 &  & 38.1 & 23.8 & 44.59\\
Code & \multirow{-3}{*}{\textbackslash{}} & 50.58 & 53.19 & 45.29 &  & 31.69 & 4.84 &  & 50.9 & 40.9 & 39.63 \\
\hline
 & \ding{55} & 47.34 & 46.80 & 41.00 &  & 52.16 & 13.26 &  & 32.1 & 29.9 & 37.51 \\
\multirow{-2}{*}{Task Arithmetic} &
  \checkmark &
  \cellcolor[HTML]{EFEFEF}\textbf{62.43} &
  \cellcolor[HTML]{EFEFEF}\textbf{61.94} &
  \cellcolor[HTML]{EFEFEF}45.29 &
  \cellcolor[HTML]{EFEFEF} &
  \cellcolor[HTML]{EFEFEF}\textbf{59.74} &
  \cellcolor[HTML]{EFEFEF}\textbf{17.06} &
  \cellcolor[HTML]{EFEFEF}  &
  \cellcolor[HTML]{EFEFEF}\underline{54.4} &
  \cellcolor[HTML]{EFEFEF}34.1 & 
  \cellcolor[HTML]{EFEFEF}\underline{47.85} \\
 & \ding{55} & 57.20 & 57.59 & \textbf{48.71} &  & 55.50 & 15.00 &  & 48.4 & 40.2 & 46.09 \\
\multirow{-2}{*}{Ties-Merging} &
  \checkmark &
  \cellcolor[HTML]{EFEFEF}57.36 &
  \cellcolor[HTML]{EFEFEF}57.94 &
  \cellcolor[HTML]{EFEFEF}\underline{48.12} &
  \cellcolor[HTML]{EFEFEF} &
  \cellcolor[HTML]{EFEFEF}56.25 &
  \cellcolor[HTML]{EFEFEF}15.56 &
  \cellcolor[HTML]{EFEFEF} &
  \cellcolor[HTML]{EFEFEF}49.9 &
  \cellcolor[HTML]{EFEFEF}\underline{41.5} & 
  \cellcolor[HTML]{EFEFEF}46.66 \\
     & \ding{55}  & 55.36 & 55.77 & 42.84 &  & 57.39 & 15.00 &  & 49.4 & 39.0 & 44.97\\
\multirow{-2}{*}{DARE} & 
  \checkmark &
  \cellcolor[HTML]{EFEFEF}\underline{58.22} &
  \cellcolor[HTML]{EFEFEF}\underline{58.92} &
  \cellcolor[HTML]{EFEFEF}46.88 &
  \cellcolor[HTML]{EFEFEF} &
  \cellcolor[HTML]{EFEFEF}\underline{58.45} &
  \cellcolor[HTML]{EFEFEF}\underline{16.46} &
  \cellcolor[HTML]{EFEFEF} &
  \cellcolor[HTML]{EFEFEF}\textbf{55.1} &
  \cellcolor[HTML]{EFEFEF}\textbf{43.3} & 
  \cellcolor[HTML]{EFEFEF}\textbf{48.19} \\
  \bottomrule
\end{tabular}%
}
\label{tab:mistral-7b}
\vspace{-0.2cm}
\end{table*}

\subsection{Main Results}

\paragraph{Merging Models with Sense-Merging.} 
We first evaluate the effectiveness of our Sens-Merging method by utilizing it as a plug-and-play module to enhance existing task-vector-based baselines. Table~\ref{tab:llama2-7b} presents the performance of the baseline methods alongside their Sens-Merging enhanced counterparts across seven datasets. Specifically, when merging fine-tuned models specialized in general knowledge (Chat\footnote{huggingface.co/meta-llama/Llama-2-7b-chat-hf}), mathematical reasoning (Math\footnote{huggingface.co/TIGER-Lab/MAmmoTH-7B}), and code generation (Code\footnote{huggingface.co/mrm8488/llama-2-coder-7b}), all derived from LLaMA2-7B\footnote{huggingface.co/meta-llama/Llama-2-7b-hf}, Sens-Merging demonstrates a consistent improvement in the average performance across all domains. Specifically, when comparing the average scores of each method with and without Sens-Merging,
we find that:

\begin{table*}[t]
\caption{Performance evaluation of merged LLaMA2-13B Models (Chat, Math, Code) across 7 task-specific datasets}
\resizebox{\textwidth}{!}{%
\begin{tabular}{lcccclcclccc}
\toprule
& & \multicolumn{3}{c}{General Knowledge} &
   &
  \multicolumn{2}{c}{Mathemetical Reasoning} &
   &
  \multicolumn{2}{c}{Code Generation} &
  \multicolumn{1}{c}{} \\ \cline{3-5} \cline{7-8} \cline{10-11} 
\multirow{-2}{*}{Method} &
  \multirow{-2}{*}{\begin{tabular}[c]{@{}c@{}}Use\\ Sens\end{tabular}} &
  \multicolumn{1}{c}{MMLU} &
  \multicolumn{1}{c}{HellaSwag} &
  \multicolumn{1}{c}{TruthfulQA} &
   &
  \multicolumn{1}{c}{GSM8K} &
  \multicolumn{1}{c}{MATH} &
   &
  \multicolumn{1}{c}{MBPP} &
  \multicolumn{1}{c}{HumanEval} &
  \multicolumn{1}{c}{\multirow{-2}{*}{Average}} \\
  \hline
Chat & & 53.17 & 60.73 & 40.88 &  & 32.37 & 6.70 &  & 16.5  & 7.9 & 31.18 \\
Math & & 52.73 & 61.10 & 37.09 &  & 55.50 & 10.84 &  & 28.8 & 15.9 & 37.42 \\
Code & \multirow{-3}{*}{\textbackslash{}} & 52.65 & 60.42 & 40.64 &  & 27.29 & 5.74 &  & 21.3 & 10.4 & 31.21\\
\hline
 & \ding{55} & 52.22 & 57.52 & \textbf{41.49} &  & 49.89 & 7.32 &  & 24.1 & 9.1 & 34.52\\
\multirow{-2}{*}{Task Arithmetic} &
  \checkmark &
  \cellcolor[HTML]{EFEFEF}\textbf{55.88} &
  \cellcolor[HTML]{EFEFEF}\textbf{61.84} &
  \cellcolor[HTML]{EFEFEF}39.05 &
  \cellcolor[HTML]{EFEFEF} &
  \cellcolor[HTML]{EFEFEF}53.07 &
  \cellcolor[HTML]{EFEFEF}8.84 &
  \cellcolor[HTML]{EFEFEF}  &
  \cellcolor[HTML]{EFEFEF}\textbf{42.6} &
  \cellcolor[HTML]{EFEFEF}20.1 & 
  \cellcolor[HTML]{EFEFEF}40.20 \\
 & \ding{55} & 55.48 & 60.65 & 39.05 &  & 52.46 & \underline{9.90} &  & 40.4 & \textbf{21.3} & 39.89 \\
\multirow{-2}{*}{Ties-Merging} &
  \checkmark &
  \cellcolor[HTML]{EFEFEF}55.20 &
  \cellcolor[HTML]{EFEFEF}60.64 &
  \cellcolor[HTML]{EFEFEF}39.17 &
  \cellcolor[HTML]{EFEFEF} &
  \cellcolor[HTML]{EFEFEF}54.44 &
  \cellcolor[HTML]{EFEFEF}\textbf{10.20} &
  \cellcolor[HTML]{EFEFEF} &
  \cellcolor[HTML]{EFEFEF}\underline{41.3} &
  \cellcolor[HTML]{EFEFEF}20.6 & 
  \cellcolor[HTML]{EFEFEF}\underline{40.22} \\
     & \ding{55}  & 55.43 & 61.51 & 40.51 &  & \underline{55.19} & 9.08 &  & 39.1 & 20.1 & 40.13\\
\multirow{-2}{*}{DARE} & 
  \checkmark &
  \cellcolor[HTML]{EFEFEF}\underline{55.65} &
  \cellcolor[HTML]{EFEFEF}\underline{61.66}  &
  \cellcolor[HTML]{EFEFEF}\underline{40.64} &
  \cellcolor[HTML]{EFEFEF} &
  \cellcolor[HTML]{EFEFEF}\textbf{55.42} &
  \cellcolor[HTML]{EFEFEF}9.08 &
  \cellcolor[HTML]{EFEFEF} &
  \cellcolor[HTML]{EFEFEF}39.3 &
  \cellcolor[HTML]{EFEFEF}\underline{20.7} & 
  \cellcolor[HTML]{EFEFEF}\textbf{40.35} \\
  \bottomrule
\end{tabular}%
}
\label{tab:llama2-13b}
\end{table*}

\begin{table*}[]
\caption{Ablation studies on task-specific scaling and cross-task scaling for Task Arithmetic in LLaMA2 7B models.}
\resizebox{\textwidth}{!}{%
\begin{tabular}{lccclcclccc}
\toprule
\multirow{2}{*}{Method} &
  \multicolumn{3}{c}{General Knowledge} &
   &
  \multicolumn{2}{c}{Mathemetical Reasoning} &
   &
  \multicolumn{2}{c}{Code Generation} &
  \multirow{2}{*}{Average} \\ \cline{2-4} \cline{6-7} \cline{9-10}
& MMLU & HellaSwag & TruthfulQA &  & GSM8K & MATH &  & MBPP & HumanEval &  \\
\hline
Task Arithmetic & 41.50 & 49.63 & 37.45 & & 47.34 & 6.46 & & 13.5 & 7.3 & 29.03 \\ 
\quad + task-specific & 41.57 & 49.60 & \textbf{37.94} &  & \textbf{48.29} & \textbf{7.84} &  & 13.3 & 7.3 & 29.41 \textcolor{blue}{(+0.38)} \\
\quad + cross-task & 45.99 & 59.07 & 36.35 & & 42.00 & 7.00      & & 32.1 & 18.3 & 33.40 \textcolor{blue}{(+5.37)} \\
\quad + Sens-Merging & \textbf{46.12} & \textbf{59.10} & 36.84 &  & 42.29 & 7.12 &  & \textbf{33.1} & \textbf{18.9} & 34.78 \textcolor{blue}{(+5.75)}\\
\bottomrule
\end{tabular}%
}
\label{tab:ablation}
\vspace{-0.3cm}
\end{table*}

\textbf{(1) Superior Improvement in Task Arithmetic:} 
Task Arithmetic exhibits a particularly notable increase from an average score of 29.03 without Sens-Merging to 34.78 with Sens-Merging, achieving a 19.22\% relative improvement of 5.58 points.
As both Ties-Merging and DARE have implemented drop strategies to mitigate parameter interference, the integration of scaling coefficient adjustments through Sens-Merging does not achieve as substantial an enhancement as seen with Task Arithmetic. Nevertheless, Sens-Merging still contributes to performance improvements in these methods, with Ties-Merging increasing from an average score of 34.73 to 35.13, and DARE improving from 34.86 to 35.13. 
\textbf{(2) Domain-Specific Improvement:}
Within the general knowledge domain, Sens-Merging significantly enhances performance on both the MMLU and HellaSwag datasets across all merging methods. In mathematical reasoning, combining Sens-Merging with the Ties-Merging baseline achieves the highest scores on both GSM8K (47.69) and MATH (7.80), surpassing their respective baselines.
In code generation, Task Arithmetic shows substantial improvements, increasing from 13.5 to 33.1 on MBPP and from 7.3 to 18.9 on HumanEval. 
\textbf{(3) Enhanced Performance than Individual Fine-tuned Models:} Sens-Merging enables the combined models to achieve higher performance on general knowledge and code generation tasks, even surpassing the original code fine-tuned model.
For example, when integrating the Chat, Math, and Code models using Sens-Merging, performance on the MBPP and HumanEval datasets increases significantly. Specifically, accuracy improves from 26.8 to 32.3 on the MBPP dataset and from 12.8 to 19.5 on the HumanEval dataset.
This demonstrates that model merging can overcome the challenges associated with training a single model for complex tasks by effectively integrating capabilities from other specialized fine-tuned models. Notably, when a Code model is merged with Math and Chat models, it achieves superior performance on coding tasks compared to code-specific fine-tuning alone.

\paragraph{Using Different Model Architecture.} To verify the generalizability of our method across architectures, we conduct experiments using Mistral-7B models. 
Using task-specific models derived from the base Mistral-7B model\footnote{huggingface.co/mistralai/Mistral-7B-v0.1} - specifically Chat\footnote{huggingface.co/mistralai/Mistral-7B-Instruct-v0.1}, Math\footnote{huggingface.co/TIGER-Lab/MAmmoTH2-7B}, and Code\footnote{huggingface.co/Nondzu/Mistral-7B-codealpaca-lora} - our method demonstrates consistent performance improvements despite the architectural differences from LLaMA-based models.
As shown in Table \ref{tab:mistral-7b}, when combined with Task Arithmetic and DARE, Sens-Merging demonstrated remarkable performance gains, surpassing the original baselines by 10.34 and 3.22 points respectively across all evaluated datasets. With Task Arithmetic, our method shows impressive gains across domains: 11.58 points in general knowledge, 4.86 points in mathematical reasoning, and 8.45 points in code generation. When combined with DARE, Sens-Merging particularly excelled in code generation, achieving a 5-point improvement over the original DARE and even outperforming task-specialized fine-tuned models. This superiority is evidenced by higher scores on coding benchmarks: 55.1 versus 50.9 on MBPP and 43.3 versus 40.0 on HumanEval.

\paragraph{Scaling to Larger Model Size.} We further evaluate the scalability of our method using the LLaMA-2 13B\footnote{huggingface.co/meta-llama/Llama-2-13b-hf} models by merging Chat\footnote{huggingface.co/meta-llama/Llama-2-13b-chat-hf}, Math\footnote{huggingface.co/TIGER-Lab/MAmmoTH-13B}, and Code\footnote{huggingface.co/emre/llama-2-13b-code-chat} fine-tuned models. As presented in Table \ref{tab:llama2-13b}, our approach maintains consistent performance gains at larger scales. Sens-Merging with Task Arithmetic demonstrates particularly strong improvements, outperforming the baseline by 5.68 points across all datasets, with notably impressive gains in code generation (14.75 points). When combined with Ties-Merging, Sens-Merging excels in mathematical reasoning tasks. Specifically, it achieves a 3.77\% relative improvement (1.98 points) on the GSM8K dataset and a 3.03\% relative improvement on the MATH dataset.


\begin{table*}[]
\caption{Performance evaluation of two merged fine-tuned LLaMA2-7B models (Chat\&Math and Math\&Code) across seven task-specific datasets.}
\resizebox{\textwidth}{!}{%
\begin{tabular}{ccccccccccccc}    
\toprule
\multirow{2}{*}{\begin{tabular}[c]{@{}c@{}}Merging\\ Methods\end{tabular}} &
  \multirow{2}{*}{Models} &
  \multirow{2}{*}{\begin{tabular}[c]{@{}c@{}}Use\\ Sens\end{tabular}} &
  \multicolumn{3}{c}{General Knowledge} &
  &
  \multicolumn{2}{c}{Mathematical Reasoning} &
  &
  \multicolumn{2}{c}{Code Generation} &
  \multirow{2}{*}{Average} \\ \cline{4-6} \cline{8-9} \cline{11-12} 
 & & & MMLU & HellaSwag & TQA & & GSM8K & MATH & & MBPP & HumanEval & \\ 
  \hline
\multirow{3}{*}{/} & Chat &
  \multicolumn{1}{c}{\multirow{3}{*}{/}} & 46.38 & 57.79 & 45.17 & & 23.43 & 4.86 & & 0.3 & 0.6 & 25.5 \\ 
 & Math &
  \multicolumn{1}{c}{} & 40.05 & 56.30 & 32.56 && 48.60 & 8.50 & & 21.8 & 12.8 & 31.52\\ 
 &
  Code &
  \multicolumn{1}{c}{} & 40.76 & 57.87 & 33.17 && 7.13 & 3.62 & & 26.8 & 5.5 & 24.98\\ 
  \hline
\multirow{4}{*}{\begin{tabular}[c]{@{}c@{}}Task\\ Arithmetic\end{tabular}} &
  \multirow{2}{*}{\begin{tabular}[c]{@{}c@{}}Chat \&Math\end{tabular}} &
  \ding{55} &
  {41.36} & 49.77 & \underline{36.96} & & 45.34 & 6.96 & & 13.8 & 7.3 & 28.78\\ 
 &
  &
  $\checkmark$ &
  \cellcolor[HTML]{EFEFEF}\textbf{45.67} &
  \cellcolor[HTML]{EFEFEF}\underline{58.02} &
  \cellcolor[HTML]{EFEFEF}\textbf{37.21} &
  \cellcolor[HTML]{EFEFEF} &
  \cellcolor[HTML]{EFEFEF}\underline{45.49} &
  \cellcolor[HTML]{EFEFEF}\underline{7.14} &
  \cellcolor[HTML]{EFEFEF} &
  \cellcolor[HTML]{EFEFEF}\textbf{30.1} &
  \cellcolor[HTML]{EFEFEF}\textbf{17.7} &
  \cellcolor[HTML]{EFEFEF}\textbf{34.48} \\ 
 &
  \multirow{2}{*}{\begin{tabular}[c]{@{}c@{}}Math \&Code\end{tabular}} &
  \ding{55} & 40.54 & 56.63 & 32.80 & & \textbf{50.42} & \textbf{9.38} & &
  {22.6} & {10.4} & {31.82} \\ 
 &
  &
  $\checkmark$ &
  \cellcolor[HTML]{EFEFEF}\underline{43.67} &
  \cellcolor[HTML]{EFEFEF}\textbf{59.15} &
  \cellcolor[HTML]{EFEFEF}33.90 &
  \cellcolor[HTML]{EFEFEF} &
  \cellcolor[HTML]{EFEFEF}42.08 &
  \cellcolor[HTML]{EFEFEF}7.12 &
  \cellcolor[HTML]{EFEFEF} &
  \cellcolor[HTML]{EFEFEF}\underline{27.8} &
  \cellcolor[HTML]{EFEFEF}\underline{16.5} &
  \cellcolor[HTML]{EFEFEF}\underline{32.89} \\ 
  \hline
\multirow{4}{*}{\begin{tabular}[c]{@{}c@{}}Ties-\\ Merging\end{tabular}} &
  \multirow{2}{*}{\begin{tabular}[c]{@{}c@{}}Chat \&Math\end{tabular}} &
  \ding{55} & \underline{45.84} & 56.48 &\underline{39.17} && \underline{48.29} &7.86 && 30.1 &\underline{17.1} & \underline{34.98}\\ 
 &
  &
  $\checkmark$ &
  \cellcolor[HTML]{EFEFEF}\underline{45.86} &
  \cellcolor[HTML]{EFEFEF}56.63 &
  \cellcolor[HTML]{EFEFEF}\textbf{40.02} &
  \cellcolor[HTML]{EFEFEF} &
  \cellcolor[HTML]{EFEFEF}\textbf{48.98} &
  \cellcolor[HTML]{EFEFEF}7.92 &
  \cellcolor[HTML]{EFEFEF} &
  \cellcolor[HTML]{EFEFEF}\textbf{31.3} &
  \cellcolor[HTML]{EFEFEF}\textbf{17.7} &
  \cellcolor[HTML]{EFEFEF}\textbf{35.49}\\ 
 &
  \multirow{2}{*}{Math\&Code} &
  \ding{55} &42.52 &\underline{58.30} &36.72 &&45.11 &
 \underline{8.44} &&29.1 &14.6 & 33.54 \\ 
 &
  &
  $\checkmark$ &
  \cellcolor[HTML]{EFEFEF}42.74 &
  \cellcolor[HTML]{EFEFEF}\textbf{58.55} &
  \cellcolor[HTML]{EFEFEF}36.96 &
  \cellcolor[HTML]{EFEFEF} &
  \cellcolor[HTML]{EFEFEF}45.49 &
  \cellcolor[HTML]{EFEFEF}\textbf{8.54} &
  \cellcolor[HTML]{EFEFEF} &
  \cellcolor[HTML]{EFEFEF}\underline{30.3} &
  \cellcolor[HTML]{EFEFEF}14.8 &
  \cellcolor[HTML]{EFEFEF}33.91 \\ 
  \hline
\multirow{4}{*}{\begin{tabular}[c]{@{}c@{}}DARE\end{tabular}} &
  \multirow{2}{*}{Chat\&Math} &
  \ding{55} &\underline{46.72} &58.04 &\textbf{39.53} &&\underline{44.88} & 6.58 && \underline{30.3} &\underline{20.1} & \underline{35.16}\\ 
 &
  &
  $\checkmark$ &
  \cellcolor[HTML]{EFEFEF}\textbf{46.78} &
  \cellcolor[HTML]{EFEFEF}58.12 &
  \cellcolor[HTML]{EFEFEF}\textbf{39.53} &
  \cellcolor[HTML]{EFEFEF} &
  \cellcolor[HTML]{EFEFEF}\textbf{45.03} &
  \cellcolor[HTML]{EFEFEF}\textbf{7.06} &
  \cellcolor[HTML]{EFEFEF} &
  \cellcolor[HTML]{EFEFEF}\underline{31.6} &
  \cellcolor[HTML]{EFEFEF}\textbf{20.7} &
  \cellcolor[HTML]{EFEFEF}\textbf{35.55}\\ 
 &
  \multirow{2}{*}{Math\&Code} &
  \ding{55} &43.95 &\underline{59.21} &33.78 &&39.80 &6.34 &&29.6 &17.1 & 32.83\\ 
 &
  &
  $\checkmark$ & \cellcolor[HTML]{EFEFEF}43.98 &
  \cellcolor[HTML]{EFEFEF}\textbf{59.25} &
  \cellcolor[HTML]{EFEFEF}33.90 &
  \cellcolor[HTML]{EFEFEF} &
  \cellcolor[HTML]{EFEFEF}40.41 &
  \cellcolor[HTML]{EFEFEF}\underline{6.82} &
  \cellcolor[HTML]{EFEFEF} &
  \cellcolor[HTML]{EFEFEF}29.8 &
  \cellcolor[HTML]{EFEFEF}17.1 &
  \cellcolor[HTML]{EFEFEF}33.04 \\ 
  \bottomrule
\end{tabular}%
}
\label{tab:2-models}
\end{table*}

\subsection{Ablation Studies}
To understand the contribution of each component in our framework, we conduct ablation studies by incorporating either task-specific scaling factors or cross-task scaling factors into the Task Arithmetic method. As shown in Table~\ref{tab:ablation}, different scaling approaches exhibit task-dependent effectiveness. For mathematical reasoning, task-specific sensitivity scaling yields notable gains (a 21.36\% relative improvement and an increase of 1.38 points on the MATH dataset) while having limited impact on other tasks. Conversely, cross-task scaling delivers significant improvements in general knowledge and code generation tasks (4.28 and 14.8 points respectively) but decreases mathematical reasoning performance by 4.8 points. Overall, cross-task scaling provides stronger aggregate performance enhancements, achieving a total gain of 5.37 points. 
Therefore, each scaling method involves a trade-off: task-specific scaling excels at enhancing specialized capabilities (particularly mathematical reasoning) but with limited broader impact, while cross-task scaling offers stronger overall performance improvements at the cost of sacrificing some task-specific excellence.

\section{In-depth Analysis}
\subsection{Scaling Factors Analysis}
\paragraph{Layerwise Sensitivity Distribution.}

\begin{figure}[t]
    \centering
    \includegraphics[width=1.0\linewidth]{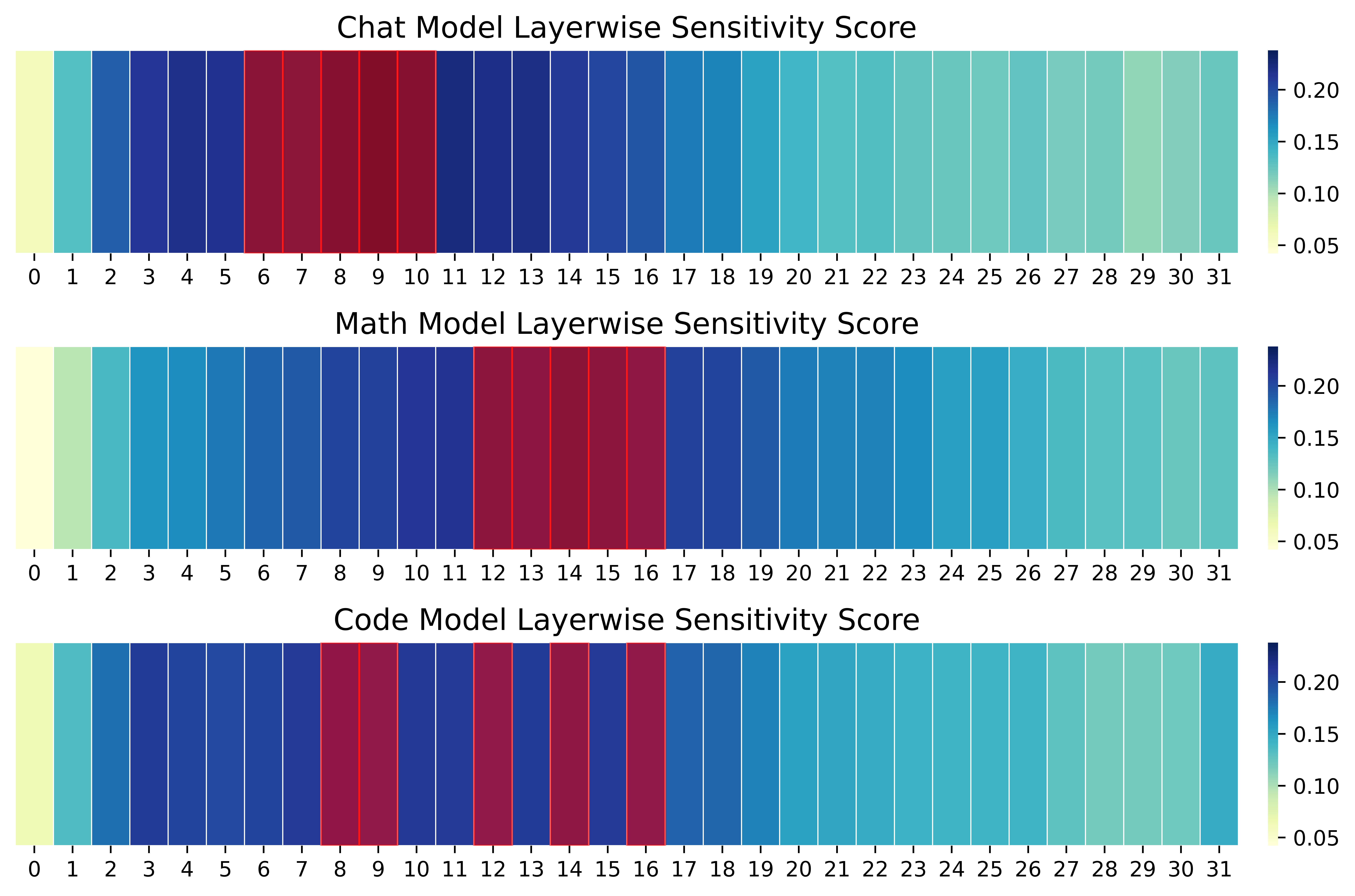}
    \caption{Layer-wise sensitivity scores across different task-specific models, with the Top-5 most sensitive layers highlighted in red.}
    \label{fig:layer_sensitivity}
    \vspace{-0.5cm}
\end{figure}

Figure~\ref{fig:layer_sensitivity} reveals distinct layer-wise sensitivity patterns across model specializations: the chat model peaks at layer 10, leveraging lower layers for language processing; the math model shows maximum sensitivity around layer 15, emphasizing middle layers for mathematical reasoning; and the code model exhibits a unique dual-peak pattern, reflecting its need for both linguistic processing in lower layers and logical reasoning in middle layers. 
Thus, by leveraging layer-wise sensitivity, we enhance the weights of the layers that are most critical to performance, thereby ensuring that specialized functions are optimally preserved.

\begin{table}[htbp]
    \centering
    \small
    \caption{Cross-task scaling coefficients across Chat, Math, and Code models.}
    \resizebox{\linewidth}{!}{%
    \begin{tabular}{lccc}
        \toprule
        Models & LLaMA2 7B & Mistral 7B & LLaMA2 13B \\ \hline
        Chat    & 0.3095   & 0.2454   & 0.2958   \\ 
        \rowcolor{gray!10}
        Math    & \textbf{0.5062}   & \textbf{0.6379}   & \textbf{0.5284}   \\ 
        Code    & 0.1843   & 0.1167   & 0.1758   \\ \bottomrule
    \end{tabular}
    }
    \label{tab:scaling}
\end{table}

\paragraph{Cross-Task Sensitivity Scaling.}

In Table \ref{tab:scaling}, we observe consistent variations in cross-task scaling factors across task vectors. 
Math models show the highest scaling coefficients (0.51-0.64), followed by Chat models (0.25-0.31), and Code models (0.12-0.18). These coefficients reveal that: mathematical reasoning provides strong transferable skills across tasks, chat abilities facilitate general language understanding, while coding skills are more specialized and less transferable.

\subsection{Merge Two Fine-tuned Models}

In addition to merging three models, we also evaluate the performance of combining two models: Chat \& Math and Math \& Code. We exclude the Chat \& Code combination as its performance is significantly lower than that of the three-model merging. 
As shown in Table \ref{tab:2-models}, when applied to two-model combinations, our Sens-merging method also outperforms baseline approaches, showing substantial improvements in Task Arithmetic method (3 points). In code generation, Sens-merging significantly improved performance over existing methods, achieving relative gains of 3.99\% (1.2 points) over Ties-Merging and 4.29\% (1.3 points) over DARE on the MBPP dataset.  
Notably, for both Ties-Merging and DARE methods, combining Chat and Math models yields better performance than merging all three models across all tasks, with Ties-Merging showing scores of 34.98 versus 34.73, and DARE showing 35.16 versus 34.86.
This indicates that simply adding more models does not guarantee better performance in merging methods. 

\section{Conclusion}

We introduce Sens-Merging, a novel method that determines model merging coefficients by analyzing parameter sensitivity both within specific tasks and across multiple tasks. Through extensive evaluation on Mistral 7B and LLaMA-2 7B/13B model families, we demonstrate that Sens-Merging enhances model merging performance across multiple domains, consistently outperforming both existing merging techniques and individually fine-tuned models. This improvement is particularly pronounced in code generation tasks, where merged models achieve superior results compared to specialized fine-tuning.




\section*{Limitations}

While Sens-Merging demonstrates remarkable performance in model merging, achieving consistent improvements across various benchmarks, it shares fundamental limitations with existing task arithmetic-based methods. For example, our current implementation primarily addresses homogeneous model merging where base models share identical architectures. While this focus allows us to achieve state-of-the-art results in such scenarios, extending Sens-Merging to heterogeneous architectures remains an exciting direction for future research. Moreover, our method is specifically designed for large language models and has been primarily validated with LoRA fine-tuned models, where weight differences between specialized models are relatively constrained. For smaller-scale models or fully fine-tuned models with larger weight divergences, our approach may require adaptations. 




\section*{Ethics Statement}
This study utilizes publicly available datasets for our models. Prior research endeavors have generally taken ethical considerations into account. We have manually inspected a subset of samples and found no explicit ethical concerns, including violent or offensive content. Nonetheless, it is crucial to highlight that the output generated by large language models lacks the degree of control we might assume. Consequently, we are prepared to implement measures to mitigate any unforeseen outputs.

\bibliography{custom}




\end{document}

%% file: latex/related_work.tex
\section{Related Work}
Modeling merging \citep{merge_survey, mergeKit}, as a complementary approach to training-based methods, has the capability to integrate multiple task-specialized models into a unified one \citep{model_soup, zipit, re-basin, dare}, to improve model performance on individual tasks by merging checkpoints without requiring additional training \citep{ties_merging, task_arithmetic}, and to alleviates the issue of catastrophic forgetting \citep{mitigating_cf}.
According to whether the based models are in same architecture, the model merging methods can be divided into heterogeneous model merging and homogeneous model merging.

\paragraph{Heterogeneous Model Merging.}
A brunch of work \citep{DBLP:conf/eccv/AvrahamiLF22,DBLP:conf/icassp/NguyenNNPBH23} attempts to perform architecture transformation before merging, aiming to transform multiple models with different architectures into a unified one. However, these approaches often rely on the learning process to align the models, which can potentially degrade their performance.
Recent research in this direction often builds upon the concept of mode connectivity \citep{DBLP:conf/iclr/FreemanB17, DBLP:conf/icml/FrankleD0C20, DBLP:conf/nips/TatroCDMSL20}, which suggests the existence of a connected path between multiple local minima of models, along which the loss remains nearly constant. Furthermore, \citet{DBLP:conf/iclr/EntezariSSN22} revealed that models permuted to the same loss basin can be merged by averaging their weights. Following these intuitions, more recent works \citep{re-basin,repair,zipit} focus on permutation strategies to achieve better heterogeneous model merging.

\paragraph{Homogeneous Model Merging.}
Task-specific models initialized from the same pre-trained model can often be merged without considering permutation symmetry \citep{model_soup, task_arithmetic}. 
One of the most straightforward approaches to model merging is to directly weighted average the parameters of base models \citep{shoemake85, model_soup}. However, the performance of simple average merging is often suboptimal, as task-specific features are typically not uniformly distributed.
Task Arithmetic \citep{task_arithmetic} enhances the merging process by introducing task vectors, suggesting that simple arithmetic operations on these vectors can effectively edit models and produce a merged model. Building on the concept of task vectors, both DARE \citep{dare} and Ties \citep{ties_merging} employ pruning-then-scaling methods to merge task vectors, based on the assumption that not all parameters contribute equally to the final performance. This also aligns with our perspective. 

Another line of research on model merging leverages information derived from the activations of training data. For example, \citet{fisher} suggested a probability-space approach, which uses the Fisher information matrix to identify the importance of model parameters and proposed Fisher Merging. \citet{reg_mean} introduced RegMean, a data-less merging method that merges models in parameter space by solving a linear system constructed from data and model parameters. 
We posit that activations play a crucial role in identifying the key parameters within task vectors relevant to downstream tasks. To this end, we propose a novel sensitivity-guided activation method to facilitate more effective merging of key features.